%% file: main.tex
\documentclass[conference]{IEEEtran}
\usepackage{cite}
\usepackage{amsmath,amssymb}
\usepackage{graphicx}
\usepackage{booktabs}
\usepackage{url}
\usepackage[hidelinks]{hyperref}
\hypersetup{pdftitle={Risk-Calibrated Intent Routing for Conversational AI via Signed Lexical Evidence},pdfauthor={Yezhou Cheng, Zehua Yang, Bojun Lin},pdfkeywords={confidence estimation, conversational AI, intent detection, lexical evidence, pretrained language models, risk calibration, selective prediction}}

\newcommand{\Pp}{\mathbb{P}}
\begin{document}
\title{Risk-Calibrated Intent Routing\\for Conversational AI via Signed Lexical Evidence}
\author{
\IEEEauthorblockN{Yezhou Cheng\textsuperscript{*}}
\IEEEauthorblockA{\textit{University of Wisconsin--Madison}\\
San Jose, USA\\
cheng222@wisc.edu\\
\textsuperscript{*}Corresponding author}
\and
\IEEEauthorblockN{Zehua Yang}
\IEEEauthorblockA{\textit{Independent Researcher}\\
San Jose, USA\\
bobbyyang13@gmail.com}
\and
\IEEEauthorblockN{Bojun Lin}
\IEEEauthorblockA{\textit{Pinterest Inc.}\\
San Jose, USA\\
bojunlin157@gmail.com}
}
\maketitle
\begin{abstract}
Conversational AI has become widely used as advances in language models let people find information, get writing help, and ask for practical guidance in their own words. When an assistant handles specific tasks, it must infer what the user wants and decide whether to act on that prediction. A sentence classifier's confidence can miss clues in the wording. We combine the gap between the classifier's two highest scores with signed lexical evidence, a score based on words and short character patterns. The score is positive when those patterns favor the predicted intent and negative when they favor another. We use separate calibration data and a statistical upper bound on the error rate to set the minimum confidence required to accept a request. Across ten runs per dataset, our method reduces the area under the risk--coverage curve, which summarizes error as more requests are accepted, by 15.8\%, 15.1\%, and 11.8\% on BANKING77, CLINC150, and HWU64, respectively, compared with a confidence model that uses only sentence classifier scores. At a 5\% error target, the share of accepted requests rises by 1.83 percentage points on BANKING77 and 5.14 points on HWU64. CLINC150 already has near-full coverage at that target. At a 2\% target, calibration finds a rule that accepts some requests in all 30 runs. Our contribution is a two-feature confidence model that improves routing by measuring how strongly the wording supports or challenges the original intent prediction.
\end{abstract}

\section{Introduction}
Advances in language models have made conversation a practical way to find information, draft text, and ask for advice. People can do all three in the same chat interface by asking in their own words. By July 2025, ChatGPT had more than 700 million weekly active users, and these activities accounted for most messages~\cite{chatterji2025usage}. Task-oriented assistants use natural language for specific tasks, such as replacing a lost bank card or setting an alarm~\cite{banking,clinc,hwu}.

To route a request, the assistant predicts the user's \emph{intent}, the task the user wants to accomplish, and selects the corresponding workflow. In banking, the phrase \emph{card payment} can appear in a request about a declined payment or an unexpected fee, each of which calls for a different response. If the prediction is uncertain, the assistant can defer the request and ask a follow-up question. This approach, known as \emph{selective routing}, aims to accept as many requests as possible for a chosen error target~\cite{selective,nlpselect}. We call the share of requests accepted \emph{coverage} and the error rate among accepted requests \emph{risk}.

One common confidence measure is the \emph{semantic margin}, the gap between the sentence classifier's two highest scores. A large gap means the model strongly prefers one intent. The wording of a request can offer a separate clue. For example, \emph{fee} may point to a payment-fee inquiry even when the sentence classifier predicts a declined payment. This motivates measuring how strongly the wording supports the specific intent that the sentence classifier chose.

We train a separate classifier that scores intents using words and short character patterns. Its \emph{signed lexical evidence} is the score for the sentence classifier's chosen intent minus the highest score for any competing intent. A positive value supports the chosen intent, and a negative value shows how strongly the wording favors a competitor. A logistic regression model, which we call the \emph{gate}, combines this evidence with the semantic margin into a confidence score. On a separate set of labeled requests, we use binomial bounds to account for uncertainty in the error rate at each candidate threshold. We choose the threshold that accepts the most requests while meeting the risk target.

We evaluate the gate over ten runs each on BANKING77, CLINC150, and HWU64. Compared with a gate that uses only sentence classifier scores, the proposed gate reduces the area under the risk--coverage curve (AURC) by 15.8\%, 15.1\%, and 11.8\%, respectively. This measure summarizes the error among accepted requests as coverage varies. At a 5\% error target, the proposed gate raises coverage by 1.83 percentage points on BANKING77 and 5.14 points on HWU64. Coverage on CLINC150 is already close to 100\%. At a 2\% target, calibration finds a rule that accepts some requests in all 30 runs at the available calibration sizes.

With the same number of features, the signed gate ranks errors better than a gate that uses the lexical classifier's confidence in its own top choice. Gains over a simple agreement indicator vary by dataset. We also test smaller calibration sets and changes in the wording and types of requests. Our contribution is a two-feature gate that uses the direction and strength of word-level evidence to improve routing while keeping intent predictions fixed.

\begin{figure*}[t]
\centering\includegraphics[width=\textwidth]{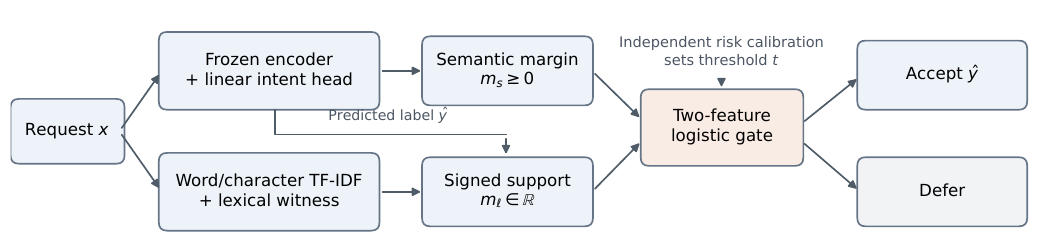}
\caption{The signed lexical gate. The sentence head supplies the label $\hat y$; the lexical witness supplies evidence for that same label. The gate changes acceptance, while the predicted intent remains fixed. Model and gate fitting precede independent threshold calibration.}
\label{fig:framework}
\end{figure*}

\section{Related Work}
\textbf{Efficient intent classification.}
Sentence encoders support efficient intent classification~\cite{banking,sbert}. CLINC150 includes explicitly out-of-scope requests~\cite{clinc}, while HWU covers conversational-agent domains~\cite{hwu}. Beyond English intent benchmarks, supervised classifiers and multilingual Transformer representations have also been compared for low-resource Urdu sentiment classification~\cite{urdusentiment}. A review of machine-learning research on fake-news detection likewise compares classical, recurrent, and Transformer classifiers and identifies cross-lingual limitations~\cite{fakenewssurvey}. We study the selective reliability of a frozen MiniLM sentence encoder~\cite{encoder} and trained linear head. Holding this predictor fixed isolates the effect of confidence estimation on routing behavior; our experiments remain limited to English intent labels.

\textbf{Confidence and complementary evidence.}
Geifman and El-Yaniv~\cite{selective} study deep selective classification with user-specified risk, and Xin et al.~\cite{nlpselect} examine confidence estimators and error regularization for NLP. Trust Score~\cite{trust} uses a density-filtered nearest-neighbor classifier to complement a model's own confidence. Liang et al.~\cite{margin} investigate margin scores under shifts, while Heng and Soh~\cite{likelihood} combine correctness-based distance scores with logit confidence. Complementary predictors and auxiliary signals also appear in other domains: Ren et al.~\cite{sensorfusion} fuse recurrent sensor models and estimate predictive variance; Wang and Yi~\cite{spatiotemporalfusion} aggregate multidimensional temporal features for video action recognition; and Ma et al.~\cite{driverdigitaltwin} combine temporal modeling with auxiliary emotion recognition for distracted-driving detection. Unlike these prediction-level fusions, our auxiliary representation refines confidence without altering the base label. These studies motivate confidence signals beyond a model's native output. Our approach contributes a complementary sparse discriminative representation, summarized by a single score aligned with the base model's selected label. Our contribution is an empirical construction for fixed intent classification using established confidence-learning and calibration components. Unsigned-confidence and agreement controls assess this construction; they do not separately identify the effects of sign, label alignment, and magnitude.

\textbf{Risk calibration and routing.}
Learn then Test (LTT)~\cite{ltt} casts risk calibration as multiple testing, and Uddin and Bauer~\cite{routing} apply a learned gate with binomial calibration to LLM routing. Selective conformal prediction sets~\cite{scrc} and e-value constructions~\cite{score} address broader selective-risk formulations. Recent agent-oriented systems apply calibrated, adversary-aware triage to reward-hacking trajectories~\cite{liu2026benchguard}, while RAGScope uses leakage-controlled, cost-aware evidence gates to route generated answers for acceptance or review~\cite{liu2026ragscope}. In coding assistants, SemVerBench shows how routing version-constraint checks to a dedicated resolver can prevent errors~\cite{chen2026semverbench}. LWCal studies the distinct problem of post-hoc probability calibration when calibration labels are noisy~\cite{liu2026lwcal}, underscoring the importance of calibration-data quality. More broadly, recommender systems map heterogeneous contextual inputs into downstream directives and face robustness, cold-start, explainability, and privacy challenges~\cite{recommendersurvey}. We use an established LTT-style procedure to translate confidence ranking into an accept-or-defer policy for binary intent errors, and compare it with a fixed-sequence alternative. This separation lets the study examine both the quality of the proposed confidence signal and the effect of the calibration procedure used to deploy it.

\section{Signed Lexical Gating}
\subsection{Prediction and witness evidence}
Let $x$ be an utterance, $y\in\{1,\ldots,K\}$ its intent, and $h(x)\in\mathbb{R}^{384}$ a normalized frozen sentence embedding. A multinomial logistic head produces logits $z(x)$ and prediction $\hat y(x)=\arg\max_k z_k(x)$. Define the semantic margin
\begin{equation}
 m_s(x)=z_{\hat y}(x)-\max_{k\ne\hat y}z_k(x).
\end{equation}
A separate linear support-vector classifier takes concatenated word and character TF-IDF features $\phi(x)$ and returns decision scores $w(x)$. Its signed support is
\begin{equation}
 m_\ell(x)=w_{\hat y}(x)-\max_{k\ne\hat y}w_k(x).
 \label{eq:signed}
\end{equation}
The index in both equations is the sentence head's prediction. Thus $m_s\geq0$, whereas $m_\ell$ can be negative. Lexical preference for a competing intent becomes negative evidence, making the score explicitly directional with respect to the routed label. Both models are fit using the same model-training partition and provide complementary views of each request.

\subsection{Two-feature correctness gate}
On a disjoint gate-fitting partition, we compute $u(x)=[m_s(x),m_\ell(x)]$ and the correctness target $b=\mathbb{1}\{\hat y=y\}$. After feature-wise standardization using that partition's means $\mu_j$ and standard deviations $\sigma_j$, a regularized logistic model yields
\begin{equation}
 s(x)=\operatorname{sigmoid}\!\left(\beta_0+
 \sum_{j\in\{s,\ell\}}\beta_j\frac{m_j(x)-\mu_j}{\sigma_j}\right).
 \label{eq:gate}
\end{equation}
The final router accepts $\hat y$ if $s(x)\geq t$ and otherwise defers. Figure~\ref{fig:framework} shows the information flow. The score ranks requests, and the next stage calibrates an operating threshold from this ranking. The gate has two coefficients and an intercept; standardization handles the different scales of logistic-head and support-vector decision scores. A hard agreement rule uses the sign of $m_\ell$, while Eq.~\eqref{eq:gate} also captures its magnitude. Inference preserves the trained models and the base classifier's labels.

\subsection{Independent risk calibration}
For a fixed threshold $t$, define coverage $C(t)=\Pp\{s(X)\geq t\}$ and selective error
\begin{equation}
 R(t)=\Pp\{\hat y(X)\ne Y\mid s(X)\geq t\}.
\end{equation}
We seek high coverage subject to a risk target $\alpha$. A second, independent development partition supplies 50 score quantiles, from quantile 0 to 0.98 in steps of 0.02; adding an accept-all threshold gives at most $M=51$ candidate rules. Deduplicate coincident thresholds. This grid is fixed before accessing risk-calibration labels.

On $n$ calibration requests, let $N_t$ be the accepted count and $K_t$ the accepted error count. With confidence-failure budget $\delta$, use the one-sided Clopper--Pearson upper bound
\begin{equation}
 U_t=\begin{cases}
 B^{-1}_{1-\delta/M}(K_t+1,N_t-K_t),&K_t<N_t,\\
 1,&K_t=N_t,
 \end{cases}
 \label{eq:cp}
\end{equation}
where $B^{-1}_q(a,b)$ is a beta quantile; $N_t=0$ also gives $U_t=1$. Select the candidate with largest $N_t$ among those satisfying $U_t\leq\alpha$. Ties favor the smaller threshold. If none qualifies, defer all requests. Hard agreement is represented by a permanent veto on disagreements, including at the accept-all candidate.

\textbf{Calibration guarantee.} Conditional on the independently fitted models, gate, and grid, assume calibration and future requests are independent and identically distributed. For a fixed candidate, conditional on $N_t$, the count $K_t$ is binomial with error probability $R(t)$. Each bound in Eq.~\eqref{eq:cp} has failure probability at most $\delta/M$. A union bound therefore gives
\begin{equation}
 \Pp_{\mathrm{cal}}\{R(\hat t)>\alpha\}\leq\delta,
 \label{eq:guarantee}
\end{equation}
with the defer-all policy assigned risk zero. This standard simultaneous-testing argument~\cite{selective,ltt} applies to each deployed policy without requiring monotonic conditional error across thresholds. The bound concerns population selective error, with probability over IID calibration samples conditional on the fitted models, gate, and grid. It is not a hard bound on a finite test set, and exchangeability alone does not justify the binomial model. Source-to-test and input-shift analyses examine practical transfer limitations without establishing the IID assumption.

\section{Experimental Design}
\subsection{Data and experimental protocol}
We use original BANKING77 and CLINC150 releases and the first HWU cross-validation fold converted by the pinned DialoGLUE script~\cite{conversion}. Table~\ref{tab:data} reports the data used. Normalization for duplicate detection lowercases and collapses whitespace; four repeated BANKING77 training texts and three HWU training texts are removed before splitting. The official test sets are retained. Seven BANKING77 and two CLINC150 test requests match normalized training texts; a sensitivity analysis excludes these overlaps and repeated test texts.

\begin{table}[t]
\caption{Dataset sizes after training deduplication. The official CLINC150 validation split is not used. HWU64 denotes the 64-intent first-fold subset.}
\label{tab:data}\centering
\begin{tabular}{lrrr}\toprule
Dataset & Intents & Train pool & Test \\\midrule
BANKING77 & 77 & 9,999 & 3,080 \\
CLINC150 & 150 & 15,000 & 4,500 \\
HWU64 & 64 & 9,957 & 1,076 \\\bottomrule
\end{tabular}
\end{table}

For each seed 101--110, the training pool is randomly divided without stratification into model fitting (60\%), gate fitting (10\%), threshold-grid development (10\%), and calibration (20\%). Calibration sizes are 2,000, 3,000, and 1,992, respectively. All methods share these partitions. A training-only seed-0 pilot selected the two-feature gate over a seven-feature variant for simplicity. Features, regularization, seeds, and primary endpoints were then fixed for the ten official evaluation runs. The primary analysis calibrates on the training-pool partition and evaluates on the official test set; we call this setting \emph{source calibration}. Target-matched calibration, calibration-procedure sensitivities, and the additional matched controls were added after the primary design freeze. They reuse frozen predictors and gates where applicable; newly fitted control gates use the original gate-fitting partition. These analyses are post-freeze sensitivity evidence, not fresh unseen-benchmark confirmation.

The encoder is \texttt{all-MiniLM-L6-v2}, revision \texttt{1110a243}, with 384-dimensional normalized embeddings and maximum sequence length 256~\cite{encoder}. The semantic head uses multinomial logistic regression with $C=10$. The lexical witness concatenates word unigrams/bigrams and within-word character 3--5-grams, with sublinear term frequency, minimum document frequency 2, separate $\ell_2$ normalization, and respective vocabulary caps of 30,000 and 50,000. A linear SVM uses $C=1$. Correctness gates use standardized inputs and logistic regression with $C=1$. No encoder weights are updated. Encoder pretraining is external to the study.

\subsection{Comparators and measurements}
\textbf{Fixed-prediction comparators} retain the semantic label: maximum softmax probability (MSP), negative entropy, raw logit margin, and a \emph{semantic gate} learned from margin, maximum probability, negative entropy, and maximum logit. Trust Score uses normalized embedding distances, ten-neighbor density filtering, and retention of the densest 90\% of examples per class~\cite{trust}. Hard agreement vetoes lexical disagreements and thresholds semantic margin. The \emph{agreement gate} adds an agreement bit to the four semantic features. A \emph{seven-feature gate} further adds signed support and the lexical model's own top-two margin. Additional saved comparators use lexical support alone or a correctness gate on embeddings plus semantic features ($C=0.1$). The latter is regularized more strongly due to its larger input dimension.

\textbf{Matched controls.} Additional two-feature gates replace signed support with either the lexical model's own unsigned top-two margin or its agreement bit. Identical fitting and calibration compare the specific signed feature with auxiliary confidence and agreement. Replacing it by the witness's own top-two margin changes both sign information and the label whose margin is measured. A $\Delta$-KNN--margin comparator~\cite{likelihood} uses correct/incorrect gate-fitting embeddings as reference sets and selects $k\in\{1,5,10,25,50\}$ (when feasible) and margin weight $\lambda\in\{0,0.01,0.1,1,10,100\}$ by grid-development AURC, before risk calibration. It uses mean log-distances to the $k$ nearest references. Full supplementary results also include unoptimized equal-probability fusion, which changes the predicted labels.

The primary endpoint is coverage under source calibration at $\alpha=0.05$, $\delta=0.05$, compared with the semantic gate. Secondary targets are 0.01, 0.02, and 0.10. We report mean coverage and conditional test error across ten splits, including zero coverage for defer-all runs; error means omit undefined empty selections. Area under the risk--coverage curve (AURC; lower is better) is the mean cumulative error after sorting test examples from highest to lowest score, and assesses ranking without choosing a calibration threshold. Paired 95\% percentile intervals for coverage differences use 5,000 crossed bootstrap draws, resampling both seeds and shared test-row indices. Additional AURC, ablation, and target-matched intervals use 2,000 paired crossed draws. All intervals are pointwise and conditional on the benchmark and fitted runs; test examples are shared across seeds. Target-matched intervals hold calibrated thresholds fixed and do not capture full uncertainty from newly sampled calibration datasets or domains.

\begin{table*}[t]
\caption{Primary source-calibrated results at $\alpha=5\%$, $\delta=5\%$. Mean accepted coverage (Cov.) and conditional test error (Err.) are percentages over ten runs; error omits undefined empty selections. Bold marks the proposed gate; italics mark the strongest non-proposed gate by mean coverage. All rows retain the same semantic predictions.}
\label{tab:main}\centering\input{generated/main_table.tex}
\end{table*}

\section{Results and Analysis}
\subsection{Error ranking and calibrated coverage}
\begin{figure*}[t]
\centering\includegraphics[width=\textwidth]{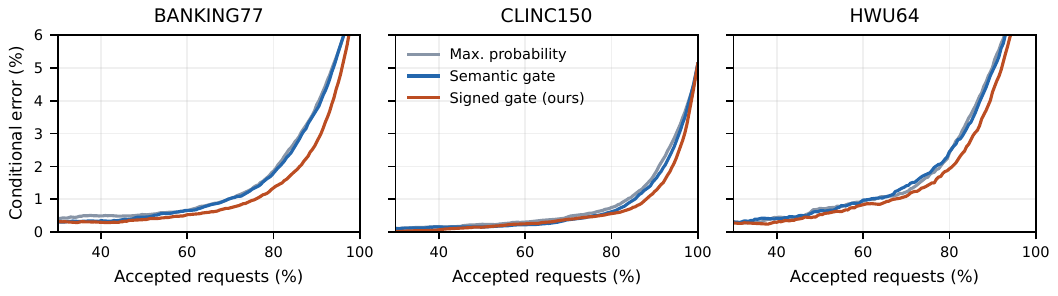}
\caption{Mean test risk--coverage curves over ten runs, retaining the same semantic predictions. Curves sweep all test score ranks for ranking analysis; operational thresholds are selected on independent calibration data. The plotted error axis is restricted to 0--6\% to show the selective operating region.}
\label{fig:rc}
\end{figure*}

The signed gate improves error ranking across all three datasets (Fig.~\ref{fig:rc}, Table~\ref{tab:aurc}). Relative to the semantic gate, AURC decreases by 15.8\%, 15.1\%, and 11.8\% on BANKING77, CLINC150, and HWU64. Paired 95\% intervals for signed-minus-semantic AURC ($\times10^3$) are $[-2.88,-1.01]$, $[-1.38,-0.28]$, and $[-3.57,-0.28]$, respectively. The lower AURC indicates improved average error ranking over coverage levels, independently of any single operational threshold; it does not establish lower risk at every coverage level.

At each signed-gate coverage, the semantic ranking yields mean errors of 3.66/4.98/3.31\%, versus 2.65/4.94/2.73\% for signed on BANKING77/CLINC150/HWU64: reductions of 1.01/0.04/0.58 points. This post hoc diagnostic uses test labels only for evaluation, not threshold selection.

At the primary 5\% calibration target, this ranking advantage translates into greater coverage on the two datasets with room to improve (Table~\ref{tab:main}). The signed gate raises mean coverage from 87.80\% to 89.63\% on BANKING77 and from 79.16\% to 84.30\% on HWU64. Paired gains are 1.83 percentage points (95\% interval 0.58--3.19) and 5.14 points (3.72--6.68), with corresponding mean errors of 2.65\% and 2.73\%. Gains over maximum softmax probability are 3.06 and 6.03 points. Thus, at the same nominal target, complementary lexical evidence enables the router to handle more requests while retaining the semantic classifier's original labels.

CLINC150 provides a complementary operating regime in which most methods already approach full acceptance at the 5\% target. In this saturated region, the signed gate contributes a 15.1\% AURC reduction alongside near-complete source-calibrated coverage. The unselected calibration/test errors are 8.50/7.99\% on BANKING77, 3.32/5.12\% on CLINC150, and 10.10/9.59\% on HWU64. The CLINC150 gap suggests calibration-transfer concerns, but does not by itself establish a population distribution difference. Signed-gate mean test error is 4.94\%, although eight of ten runs exceed 5\%; one run has 3.06\% error. These shared-test runs are not independent trials of the certificate's failure probability. Mean error decreases to 2.98\% in the target-matched sensitivity of Section~V-D, whose calibration budget and evaluation subset also differ.

\begin{figure*}[t]
\centering\includegraphics[width=\textwidth]{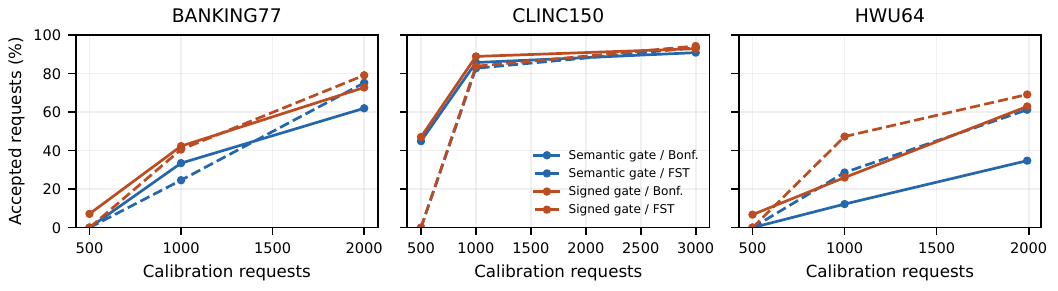}
\caption{Calibration-budget sensitivity at a 2\% error target. Solid lines use simultaneous Bonferroni bounds; dashed lines use fixed-sequence testing (FST). Coverage includes defer-all outcomes and is averaged over ten runs. The full calibration sizes differ by dataset. Models, gates, and independent grids stay fixed.}
\label{fig:cal}
\end{figure*}

\begin{table}[t]
\caption{Mean AURC $\times 10^3$; lower is better. Two-feature agreement and unsigned gates match the proposed gate's dimension. All rows use the same base predictions.}
\label{tab:aurc}\centering\input{generated/aurc_table.tex}
\end{table}

\subsection{Why signed evidence matters}
Matched controls support the effectiveness of the proposed prediction-aligned feature relative to the tested unsigned feature. The unsigned gate has AURC 12.11/5.03/15.13, compared with 10.44/4.51/13.64 for signed support (Table~\ref{tab:aurc}). Paired signed-minus-unsigned intervals ($\times10^3$) are $[-2.21,-1.16]$, $[-0.88,-0.20]$, and $[-2.60,-0.52]$. This comparison favors the proposed feature, but does not isolate the separate effects of sign and label alignment. The two-feature agreement gate provides a strong matched comparison: signed support adds 0.36 coverage points on BANKING77 and 2.37 points on HWU64 (95\% interval 1.38--3.41), with the BANKING77 interval crossing zero. Signed-minus-agreement AURC intervals ($\times10^3$) are $[-1.52,-0.08]$, $[-0.82,0.09]$, and $[-2.25,0.37]$. Thus, the incremental benefit over binary agreement varies by dataset and metric; the evidence does not establish a universal benefit from continuous magnitude alone.

The remaining controls clarify how the two signals complement one another. Lexical support alone gives AURC 12.40/6.90/16.87, supporting its combination with semantic margin. The seven-feature gate changes coverage by less than 0.4 points in either direction, showing that the compact construction captures most of the available benefit. Although learned without a sign constraint, standardized signed-support coefficients are positive in all 30 fitted gates (range 0.89--1.71). The $\Delta$-KNN--margin comparator yields primary coverage 87.52/99.78/78.30\%; signed-gate gains on BANKING77 and HWU64 are 2.11 and 6.00 points, while CLINC150 remains saturated. AURC intervals support the signed gate's advantage on BANKING77 and CLINC150; the HWU64 interval spans zero.

The continuous score also allows the gate to trade evidence across agreement regions instead of imposing a categorical veto. At the primary target, it accepts lexical-disagreement requests amounting to 2.49\% of all BANKING77 requests and 2.04\% of HWU64 requests; these subsets have mean errors of 14.33\% and 6.50\%, respectively. The deployed criterion controls aggregate accepted risk, while group-specific calibration offers a direct extension for applications that assign a separate risk budget to lexical disagreements.

\subsection{Strict targets and calibration budgets}
At strict targets, ranking quality and the calibration procedure jointly determine attainable coverage. At $\alpha=2\%$, the signed gate's mean source-calibrated coverage is 72.77\%, 92.93\%, and 62.94\% (Table~\ref{tab:strict}), and simultaneous binomial calibration returns a nonempty policy in all 30 dataset--seed runs at the available budgets. The semantic gate defers every request in one BANKING77 run and three HWU64 runs. Simultaneous testing sets a substantial evidence requirement when few accepted errors can be tolerated: with zero observed errors, Eq.~\eqref{eq:cp} requires at least $\lceil\log(\delta/M)/\log(1-\alpha)\rceil$ accepted calibration requests. At $M=51$, $\alpha=0.02$, and $\delta=0.05$, this is 343 requests. A stronger confidence ranking helps assemble a sufficiently large set of reliable calibration examples.

\begin{table}[t]
\caption{Mean coverage/error (\%) under strict source calibration and target-matched sensitivity analysis. Error means omit empty sets; coverage includes them. At 2\%, the semantic gate defers all in one BANKING77 and three HWU64 runs; the signed gate never does.}
\label{tab:strict}\centering\setlength{\tabcolsep}{3pt}\input{generated/strict_table.tex}
\end{table}

We further compare simultaneous calibration with LTT fixed-sequence testing~\cite{ltt}. The independently specified sequence starts at the development-set threshold for 20\% coverage and increases coverage in two-point steps, ending with accept-all. Each candidate uses a level-$\delta$ binomial bound, and testing stops at the first non-rejection. Figure~\ref{fig:cal} evaluates both procedures with 500, 1,000, and all available calibration requests. At the 2\% target on HWU64, fixed-sequence testing yields 61.25\% semantic-gate coverage and 69.15\% signed-gate coverage. The 28.14-point difference under Bonferroni calibration becomes 7.91 points under fixed-sequence testing, retaining an advantage for signed evidence. This comparison separates two contributors to deployable coverage: confidence quality and the statistical procedure used to calibrate it.

\subsection{Calibration transfer and input shift}
Calibration data determine how ranking quality translates into an operational policy. We examine source-to-target transfer by randomly splitting each official test set in half, calibrating on one half, and evaluating on the other, using seed offsets fixed at 7000. Predictors, gates, and training-derived grids remain unchanged. Table~\ref{tab:strict} reports both coverage and error under this target-matched setting. Signed-minus-semantic coverage gains are 3.92 points (95\% interval 2.85--5.04), 1.26 (0.36--2.10), and 4.78 (2.06--7.60) on BANKING77, CLINC150, and HWU64. The within-setting signed-versus-semantic advantage persists. However, source-to-target comparisons also change calibration size and evaluation rows: CLINC150 uses 2,250 target-calibration requests instead of 3,000 source requests, and evaluates on 2,250 rather than 4,500 requests. The observed error decrease is consistent with representative calibration, but cannot be attributed exclusively to source matching; calibration-budget conservatism and subset variation can also contribute. An equal-budget comparison on identical evaluation rows would isolate the source effect more directly.

An explicit covariate stress test further identifies the data needed to sustain calibrated operation. Robustness reviews in deepfake detection similarly emphasize adversarial strategies, generalization challenges, and multimodal evidence~\cite{deepfakesurvey}, although our stressors are lexical rather than visual. We deterministically swap two distinct adjacent internal letters in one eligible alphabetic word per test request, retaining the original labels as a diagnostic proxy. Without recalibration, signed-gate coverage/error becomes 74.00/6.04\%, 98.89/19.35\%, and 67.49/7.05\%, respectively. With a weighted 10\% mixture of CLINC out-of-scope requests, counting every accepted out-of-scope request as wrong, signed-gate error reaches 14.11\%. These results quantify the effect of two deployment changes---orthographic perturbations and unsupported intents---and motivate representative calibration streams together with explicit out-of-scope handling. Recalibration on representative requests is a possible response under the stated IID conditions, but these stress experiments do not test whether it restores useful coverage under either shift. They do not establish out-of-scope detection or downstream agent safety.

No universal recalibration interval follows from these experiments. Deployment therefore needs a prespecified drift or labeled-monitoring trigger and representative labels; validating such a trigger is future work.

\subsection{Reproducibility}
The archived package (\href{https://doi.org/10.5281/zenodo.23461612}{doi:10.5281/zenodo.23461612}) provides a locked environment, pinned data/model revisions, splits, per-request outputs, calibration scores, and supplementary results. An independent exact-binomial implementation reproduces all 1,800 operating points and acceptance masks; the archive also retains overlap-filtered comparisons, full 1\% and 10\% results, and unclipped risk--coverage curves. On an Apple M1 Pro with four CPU threads, seed-101 single-request measurements (200 after 20 warmups per dataset) give 6.54--6.76 ms median semantic latency plus 0.90--0.92 ms for the lexical witness and gate, excluding model loading.

\section{Discussion}
The results establish prediction-aligned lexical support as a useful complement to semantic confidence. The lexical witness focuses exclusively on evaluating how strongly a distinct representation supports the label selected by the sentence classifier. This alignment converts lexical agreement and disagreement into graded evidence and allows the gate to retain useful requests through a continuous tradeoff. Improvements over the tested unsigned feature across all three datasets support this construction, while benefits beyond binary agreement depend on the dataset and metric. The current controls do not separately identify sign, alignment, and magnitude. The close results of the larger seven-feature gate support the compact construction.

The evaluation also separates ranking quality from the calibration procedure that turns a score into a routing policy. AURC improvements are consistent across all three datasets, including CLINC150, where near-full source-calibrated acceptance leaves little headroom at the primary operating point. Under stricter targets, stronger ranking helps calibration identify nonempty policies, while the fixed-sequence comparison quantifies how the calibration procedure affects absolute coverage. The post-freeze target-matched analysis supports within-setting method comparisons, while changing calibration source, budget, and evaluation rows prevents isolating the source effect. The shift failures motivate representative calibration and explicit out-of-scope handling; recovery through recalibration remains untested here.

\subsection{Scope and future opportunities}
The controlled evaluation isolates confidence estimation on English single-intent benchmarks using one frozen sentence encoder. Future work should test other encoders, languages, and multi-intent requests~\cite{zou2022multilabel}. Mechanism-focused extensions should add absolute-aligned-margin and agreement-plus-unsigned-confidence controls to separate sign, label alignment, and magnitude. Repeated splits measure fitting variability on shared benchmark test sets; the target-matched analysis is a sensitivity, not an isolated test of calibration-source matching. Equation~\eqref{eq:guarantee} provides population aggregate risk control under conditional IID calibration and deployment requests. Group-aware calibration is needed for per-intent or lexical-disagreement budgets; the guarantee covers neither these groups nor downstream execution success. User heterogeneity may further motivate group-conditioned extensions: Du et al.~\cite{personalizedlanechange} use behavior-based personalization for lane-change prediction, while intent routing may analogously require calibration tailored to user groups. Longitudinal deployment studies should validate drift or labeled-monitoring triggers, quantify representative-label budgets and recalibration frequency, and test recovery under shift. Further work can connect these confidence policies to complete assistant workflows, out-of-scope detection, and downstream recovery.

\section{Conclusion}
This work introduces a signed lexical gate for improving the selective reliability of a fixed intent classifier. The gate aligns sparse lexical evidence with the classifier's predicted intent and combines it with semantic margin, preserving the original label while refining the decision to accept or defer. Across BANKING77, CLINC150, and HWU64, the resulting score improves error ranking relative to a learned semantic gate; at the primary calibration target, it also increases accepted coverage on the two datasets with available headroom. Matched controls favor the proposed feature over the tested unsigned auxiliary confidence, with mixed incremental benefits over binary agreement. Calibration sensitivities show that operational coverage also depends on the procedure and data budget; source-transfer and shift results bound the deployment conclusions. Signed lexical confidence therefore offers a compact and interpretable way to strengthen risk-calibrated intent routing without retraining or altering the underlying classifier.

\bibliographystyle{IEEEtran}
\bibliography{references}
\end{document}

%% file: generated/main_table.tex
\begin{tabular}{lrrrrrr}
\toprule
& \multicolumn{2}{c}{BANKING77} & \multicolumn{2}{c}{CLINC150} & \multicolumn{2}{c}{HWU64} \\
Method & Cov. & Err. & Cov. & Err. & Cov. & Err. \\
\midrule
Max. probability & 86.57 & 3.09 & 99.78 & 5.01 & 78.27 & 2.12 \\
Logit margin & 86.72 & 3.04 & 99.80 & 5.03 & 77.90 & 2.40 \\
Semantic gate & 87.80 & 3.22 & 99.79 & 5.00 & 79.16 & 2.32 \\
Trust score & 87.24 & 2.95 & 99.69 & 5.06 & 71.47 & 3.09 \\
Hard agreement & 88.03 & 2.73 & 91.18 & 1.90 & 81.91 & 2.79 \\
Agreement gate & 89.10 & 2.72 & 99.76 & 4.97 & 81.98 & 2.45 \\
Unsigned gate & 88.07 & 3.24 & 99.77 & 5.01 & 79.99 & 2.44 \\
Two-feature agreement & 89.27 & 2.77 & 99.75 & 4.96 & 81.93 & 2.44 \\
$\Delta$-KNN + margin & 87.52 & 3.11 & 99.78 & 5.01 & 78.30 & 2.20 \\
\textit{Seven-feature gate} & \textit{89.59} & \textit{2.65} & \textit{99.74} & \textit{4.95} & \textit{83.99} & \textit{2.59} \\
Lexical support & 87.29 & 2.75 & 99.68 & 4.95 & 81.69 & 3.11 \\
\textbf{Signed gate (ours)} & \textbf{89.63} & \textbf{2.65} & \textbf{99.73} & \textbf{4.94} & \textbf{84.30} & \textbf{2.73} \\
\bottomrule
\end{tabular}

%% file: generated/aurc_table.tex
\begin{tabular}{lrrr}
\toprule
Method & BANKING77 & CLINC150 & HWU64 \\
\midrule
Semantic gate & 12.41 & 5.31 & 15.47 \\
Unsigned gate & 12.11 & 5.03 & 15.13 \\
Two-feature agreement & 11.26 & 4.84 & 14.57 \\
$\Delta$-KNN + margin & 12.54 & 5.42 & 15.38 \\
Lexical support & 12.40 & 6.90 & 16.87 \\
Seven-feature gate & 10.50 & 4.49 & 13.69 \\
Signed gate & 10.44 & 4.51 & 13.64 \\
\bottomrule
\end{tabular}

%% file: generated/strict_table.tex
\begin{tabular}{lrrr}
\toprule
Gate & BANKING77 & CLINC150 & HWU64 \\
\midrule
\multicolumn{4}{l}{Source calibration, $\alpha=2\%$} \\
Max. probability & 66.2/0.97 & 90.2/1.84 & 40.0/0.70 \\
Semantic gate & 62.0/1.00 & 90.9/1.77 & 34.8/0.54 \\
Signed gate & 72.8/0.96 & 92.9/1.76 & 62.9/0.83 \\
\midrule
\multicolumn{4}{l}{Target-matched calibration, $\alpha=5\%$} \\
Max. probability & 86.2/2.91 & 95.5/3.22 & 75.6/1.80 \\
Semantic gate & 86.6/2.84 & 95.6/3.06 & 75.2/1.79 \\
Signed gate & 90.5/2.83 & 96.8/2.98 & 80.0/2.13 \\
\bottomrule
\end{tabular}